\RequirePackage[OT1]{fontenc}
\documentclass[letterpaper, 10 pt, conference]{ieeeconf}  

\IEEEoverridecommandlockouts                              

\usepackage[utf8]{inputenc}
\usepackage{graphics} 
\usepackage{epsfig} 
\usepackage{amsmath} 
\usepackage{mathtools}
\usepackage{amsfonts}
\usepackage{amssymb}  
\usepackage{multirow, hhline}
\usepackage{algorithm}
\usepackage{algpseudocode}
\usepackage{lipsum}
\usepackage{subfloat}

\usepackage[export]{adjustbox}
\usepackage{soul,color}
\usepackage{mathdots}
\usepackage{caption, subcaption}

\usepackage{textcomp}
\usepackage{gensymb}

\newcommand\ptwiddle[1]{\mathord{\mathop{#1}\limits^{\scriptscriptstyle(\sim)}}}
\allowdisplaybreaks

\makeatletter
\let\NAT@parse\undefined
\makeatother
\usepackage{hyperref} 

\title{\LARGE \bf
Fast Autonomous Robotic Exploration Using the Underlying Graph Structure
}

\author{Julio A. Placed and Jos\'e A. Castellanos
  \thanks{The authors are with the \textit{Departamento de Inform\'atica e Ingenier\'ia de Sistemas, Instituto de Investigaci\'on en Ingenier\'ia de Arag\'on, Universidad de Zaragoza}, C/ Maria de Luna 1, 50018, Zaragoza, Spain.  {\tt\small \{jplaced, jacaste\}@unizar.es}}%
}

\begin{document}

\onecolumn

\begin{center}
    This paper has been accepted for publication in \textit{2021 IEEE/RSJ International Conference on Intelligent Robots and Systems (IROS)}.
\end{center}
\bigskip
\begin{center}
    DOI: \href{https://ieeexplore.ieee.org/abstract/document/9636148}{10.1109/IROS51168.2021.9636148}
\end{center}
\bigskip
© 2021 IEEE. Personal use of this material is permitted. Permission from IEEE must be obtained for all other uses, in any current or future media, including reprinting/republishing this material for advertising or promotional purposes, creating new collective works, for resale or redistribution to servers or lists, or reuse of any copyrighted component of this work in other works.

\clearpage
\twocolumn
\maketitle
\thispagestyle{empty}
\pagestyle{empty}

\begin{abstract}

In this work, we fully define the existing relationships between traditional optimality criteria and the connectivity of the underlying pose-graph in Active SLAM, characterizing, therefore, the connection between Graph Theory and the Theory Optimal Experimental Design. We validate the proposed relationships in 2D and 3D graph SLAM datasets, showing a remarkable relaxation of the computational load when using the graph structure. Furthermore, we present a novel Active SLAM framework which outperforms traditional methods by successfully leveraging the graphical facet of the problem so as to autonomously explore an unknown environment.

\end{abstract}


\section{Introduction} \label{S:1}

Simultaneous Localization and Mapping (SLAM) describes the problem of incrementally building the map of a previously unseen environment while at the same time locating the robot on it. Since its statement, it has been a key topic in the robotics community and numerous approaches have been developed. Some of them, often based on filter techniques, aim to estimate the robot's pose solely at a certain time, and the map. In contrast, full SLAM approaches focus on estimating the entire trajectory and the map. Graph SLAM is the most common way to tackle the latter, in which the nodes of the graph encode the poses and the edges encode the constraints between them. Once the graph is built, it all comes down to finding the optimal configuration through Maximum Likelihood (ML), i.e., finding the set of poses that minimize the constraints associated to the observations. See \cite{thrun02, durrant06, grisetti10, cadena16} and references there in.

The previous problem may be augmented by also including on it the optimal choice of actions the robot should execute, which implies the optimal choice of the measurements it receives. This new problem, built on the basis of Active Localization \cite{burgard97}, is known as Active SLAM, and can be formally defined as the paradigm of controlling a robot which is performing SLAM so as to reduce the uncertainty of its localization and the map's representation \cite{feder99, carrillo12}. Typically, it is divided in three phases \cite{makarenko02}: (i) the identification of all possible locations to explore (ideally infinite), (ii) the computation of the utility associated to the actions that would take the robot from its current position to each of those locations and (iii) the selection and execution of the optimal action. During the second step, the utility of each action is computed by quantifying the uncertainty in the estimation of the target random variables. This quantification is traditionally done on the basis of either Theory of Optimal Experimental Design (TOED) or Information Theory (IT), and is just a scalar mapping of the covariance matrix \cite{pukelsheim06}. Works in \cite{leung06,carlone14bis, bai16} and \cite{carrillo18} achieve robotic exploration by using metrics that stem from the previous.

Nonetheless, the use of dense covariance matrices is computationally heavy and quickly becomes intractable in full SLAM approaches. In this sense, Khosoussi \textit{et al.} \cite{khosoussi14,khosoussi19} recently observed that the connectivity of the underlying pose-graph is closely related to the optimality criteria of Active SLAM. In particular, they showed the existing relationship between the number of spanning trees of a graph and its algebraic connectivity and the D- and E-optimality criteria, respectively, for the 2D graph SLAM case in which the covariance matrix is constant through measurements.  This idea, however, was already proposed four decades ago, when Cheng \cite{cheng81} realized that a graph with the maximum number of spanning trees is D-optimal, and related two problems that had always been viewed differently. Thus, instead of maximizing optimality criteria over the Information matrix (FIM), optimal actions in Active SLAM can be found through the maximization of the graph connectivity indices, which usually come from the analysis of the Laplacian spectrum. \cite{chen20} used the previous insights to build a 2D Active SLAM algorithm that achieves uncertainty reduction by exploiting the graphical structure.

\begin{figure}[t!]
    \centering
    \includegraphics[width=0.95\linewidth]{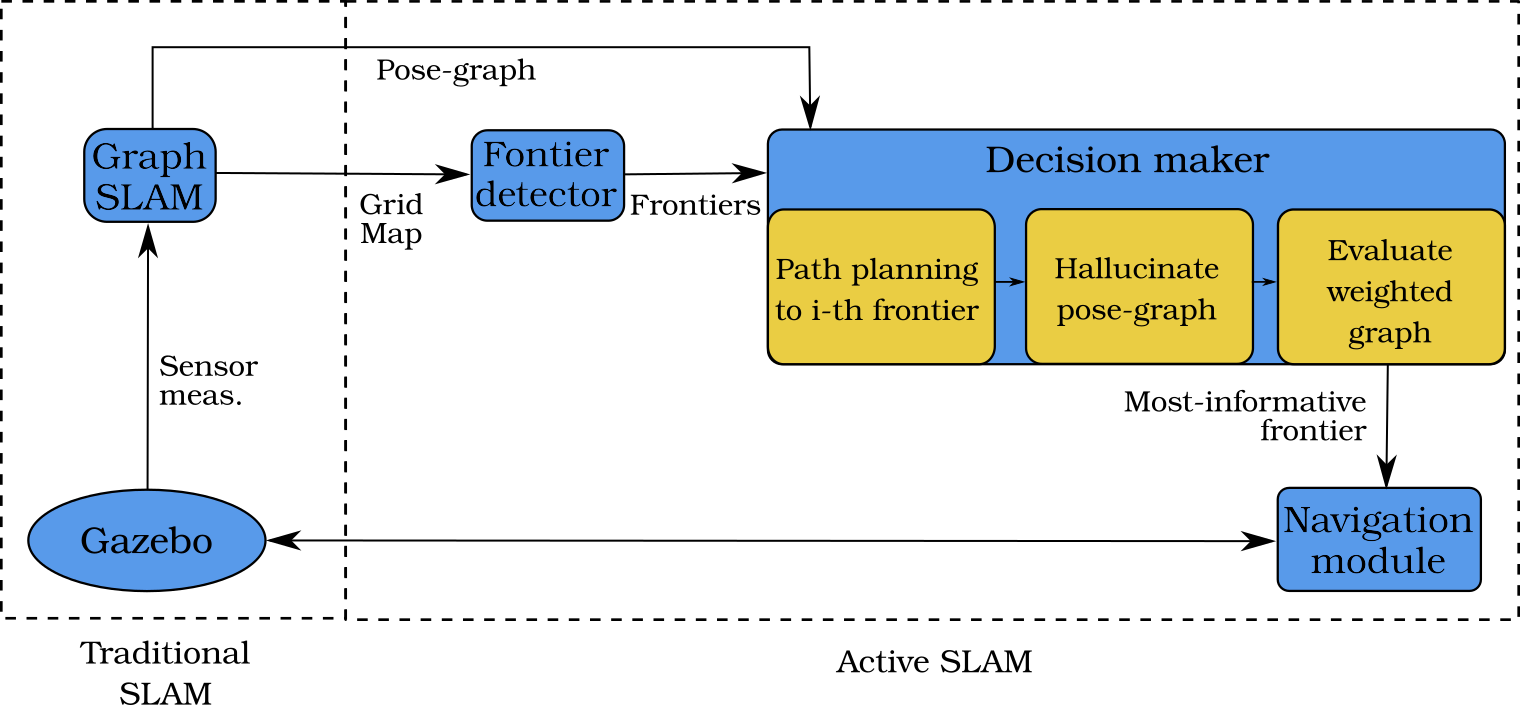}
    \caption{Overview of the proposed Active SLAM system.}
    \label{fig:active_system}
\end{figure}

In this work, on the basis of \cite{khosoussi14} and differential models, we propose a general theoretical relationship between optimality criteria of the FIM and those of the Laplacian (Section \ref{S:3}). This allows to formulate Active SLAM problem in terms of the structure of the underlying graph and to quantify utility via Graph Theory. We validate such formulation with different datasets (Section \ref{S:4}), showing that exploiting the graph structure leads to equivalent results with a remarkably lower time consumption. Moreover, in Section \ref{S:5}, we present an online Active SLAM approach based on the foregoing (see overview in Fig. \ref{fig:active_system}). Our method is capable of picking D-optimal actions in a significantly lower time than other traditional state-of-the-art methods. Thus, it can explore larger areas in a fixed time frame while maintaining low uncertainty. Finally, conclusions and future work are outlined in Section \ref{S:6}.

\section{Background}\label{S:2}

\subsection{Optimality Criteria}

As mentioned in the previous section, the utility of taking a certain action during Active SLAM is quantified by the mapping $\|\boldsymbol{\Sigma}\|\to\mathbb{R}$, being $\boldsymbol{\Sigma}\in\mathbb{R}^{\ell\times\ell}$ the positive semidefinite covariance matrix which measures the uncertainty of the $\ell$-dimensional state vector to be estimated, $(x_1,\dots,x_\ell)^T$. Kiefer \cite{kiefer74}, on the basis of TOED, showed that there is indeed a family of such mappings dependent of just one parameter ($p$); which may be expressed as follows:
\begin{equation}
    \|\boldsymbol{\Sigma}\|_p = \left\{
        \begin{array}{lrr}
            \left( \frac{1}{\ell}\sum\limits_{k=1}^\ell \lambda_k^p \right)^{\frac{1}{p}} & \text{if} &p\neq0,p\leq1 \\[1em]
            \exp \left(\frac{1}{\ell} \sum\limits_{k=1}^\ell \log(\lambda_k) \right) & \text{if} &p=0
        \end{array} \right. \label{eq:utility}
\end{equation}

In essence, utility functions are functionals of the eigenvalues of $\boldsymbol{\Sigma}$, $(\lambda_1,\dots,\lambda_\ell)$. Four optimality criteria are inferred from the previous equation:
\begin{itemize}
  \item T-optimality criterion ($p=1$): captures the average variance,
  \begin{equation}
    T{\text -} opt \triangleq \frac{1}{\ell}\sum_{k=1}^\ell \lambda_k \label{eq:topt}
  \end{equation}

  \item D-optimality criterion ($p=0$): captures the whole variance hyper-ellipsoid,
  \begin{equation}
    D{\text -} opt \triangleq \exp \left(\frac{1}{\ell} \sum_{k=1}^\ell \log(\lambda_k) \right) \label{eq:dopt}
  \end{equation}

  \item A-optimality criterion ($p=-1$): captures the harmonic mean variance,
  \begin{equation}
    A{\text -} opt \triangleq \left(\frac{1}{\ell}\sum_{k=1}^\ell \lambda_k^{-1}\right)^{-1} \label{eq:aopt}
  \end{equation}

  \item E-optimality criterion ($p\to-\infty$): captures the minimum eigenvalue,
  \begin{equation}
    E{\text -} opt \triangleq \min (\lambda_k : k=1,...,\ell) \label{eq:eopt}
  \end{equation}

\end{itemize}

\noindent Note that the maximum eigenvalue may be also captured for $p\to\infty$: $\tilde{E}{\text -} opt \triangleq \max (\lambda_k : k=1,...,\ell)$; and that the same analysis can be done on the Information matrix, $\mathbf{Y}=\boldsymbol{\Sigma}^{-1}$.

\subsection{Graph Theory Fundamentals}

Consider the pose-graph obtained from a graph SLAM algorithm to be defined by the strict directed graph $\boldsymbol{\mathcal{G}}\triangleq(\mathcal{V},\mathcal{E})$, where each vertex $\mathbf{v}_i\in\mathcal{V}$ is a robot's pose and each edge $\mathbf{e}_j\triangleq(\mathbf{v}_i,\mathbf{v}_k)\in\mathcal{E}$ represents a constraint between two poses. The dimension of the sets of vertices and edges will be denoted as $|\mathcal{V}|=n$ and $|\mathcal{E}|=m$, respectively. The simplest element used to characterize these graphs and capture their properties is the adjacency matrix, $\mathbf{A}\in\{0,1\}^{n\times n}$, a square matrix which columns and rows represent the nodes of the graph and each element $a_{i,k}$ is equal to $1$ if the nodes $v_i$ and $v_k$ are connected and $0$ otherwise. The degree matrix, $\mathbf{D}\in\mathbb{N}^{n\times n}$, is a diagonal matrix in which each element is given by $\mathrm{deg}(\mathbf{v}_i)$. The incidence matrix, $\mathbf{Q}$, shows the connections between vertices and edges. It can be defined as a concatenation of $m$ $n$-column vectors, $\mathbf{Q}=(\mathbf{q}_1, \mathbf{q}_2, \dots, \mathbf{q}_m)\in\{-1,0,1\}^{n\times m}$. Any element $[q_j]_i=-[q_j]_k=1$ if vertex $(\mathbf{v}_i,\mathbf{v}_k)$ are incident upon edge $\mathbf{e}_j$, and $0$ otherwise. Finally, the Laplacian matrix, $\mathbf{L}\in\mathbb{Z}^{n\times n}$, is a complete representation of the graph, and may be read as a particular case of the discrete Laplace operator. It can be expressed as a combination of the previously defined matrices, as $\mathbf{L}\triangleq \mathbf{D}-\mathbf{A}$, or more interestingly, as $\mathbf{L}\triangleq \mathbf{Q} \mathbf{Q}^T = \mathbf{q}_1 \mathbf{q}_1^T + ... + \mathbf{q}_m \mathbf{q}_m^T$. The latter allows to formulate the Laplacian matrix as the sum of the effects of each edge,
\begin{equation}
    \mathbf{L} \triangleq \sum_{j=1}^m \mathbf{E}_j \label{eq:laplacian}
\end{equation}
where $\mathbf{E}_j=\mathbf{q}_j \mathbf{q}_j^T\in\{-1,0,1\}^{n\times n}$ represents the connection between a pair of vertices $(\mathbf{v}_i,\mathbf{v}_k)$ through the $j$-edge. An element of the matrix diagonal is known to be $1$ if it is associated to the vertices, i.e. $[E_j]_{i,i}$ and $[E_j]_{k,k}$; and $0$ otherwise. Off-diagonal elements are $-1$ if the nodes are related, i.e. $[E_j]_{i,k}$ and $[E_j]_{k,i}$; and $0$ otherwise.

For a weighted graph $\boldsymbol{\mathcal{G}}_w$ in which edges are defined by $\tilde{\mathbf{e}}_j\triangleq(\mathbf{v}_i,\mathbf{v}_k,w_{i,k})$, generalization is straight-forward for the case $w_j\equiv w_{i,k} \in\mathbb{R}$. The weighted Laplacian is defined by:
\begin{equation}
    \mathbf{L}_w \triangleq \sum_{j=1}^m \mathbf{E}_j \ w_j \equiv \left\{
        \begin{array}{ll}
            -w_{i,k} & \text{if} \ \ \ i\neq k, \ a_{i,k}=1 \\
            0 & \text{if} \ \ \ i\neq k, \ a_{i,k}=0 \\
            \sum\limits_{q=1}^n w_{i,q} & \text{if} \ \ \ i=k
        \end{array} \right. \label{eq:wlaplacian}
\end{equation}
\noindent Note that \eqref{eq:wlaplacian} yields to \eqref{eq:laplacian} when $w_j=1\ \forall j$. Also, $\mathbf{L}_w$ is symmetric, positive semidefinite and singular, since $\mathbf{L}_w \mathbf{1}^T = \mathbf{0}^T$.

\section{A General Relationship between Information and Laplacian Matrices}\label{S:3}

Consider now a SLAM pose-graph in which each edge is weighted by the covariance matrix $\boldsymbol{\Sigma}_j \in\mathbb{R}^{\ell\times \ell}$ or, equivalently, by $\boldsymbol{\phi}_j=\boldsymbol{\Sigma}_j^{-1}$. Then, the full graph will have a FIM, $\mathbf{Y} \in \mathbb{R}^{n\times n}$, which will be a block matrix with $\ell\times \ell$ sub-matrices. This matrix is populated with zero blocks everywhere but where a pair of vertices are connected, and its dimensions will grow as the trajectory does. It can be expressed iteratively as the sum of the contributions of each edge,
\begin{equation}
    \mathbf{Y}=\sum_{j=1}^m \mathbf{Y}_j \label{eq:0}
\end{equation}
\noindent where $\mathbf{Y}_j$ is the FIM of the full graph associated to the $j$-th edge. Every block element will be zero except for those associated to the vertices related by that edge, which will be $\boldsymbol{\phi}_j$ if they are in the diagonal and $-\boldsymbol{\phi}_j$ otherwise. We will show that, under certain conditions, it is possible to relate \eqref{eq:wlaplacian} and \eqref{eq:0}.

In the general case of the SLAM problem, the robot's pose and its uncertainty can be defined with Lie groups using $SE(n)$. Just like in differential representations, the location of the robot w.r.t. a global frame, $\mathbf{T}_{wi}$, is defined by a large noise-free value which contains the estimated location, and a small (differential) perturbation which fully encodes the estimation error, that is:
\begin{equation}
    \mathbf{T}_{wi} = \bar{\mathbf{T}}_{wi} \ \exp{(\hat{\mathbf{d}}_i)}
\end{equation}
\noindent where $\bar{\mathbf{T}}_{wi}\in SE(n)$ is the estimation, and $\mathbf{d}_i\equiv \mathbf{d}_{ii}$ is a random vector normally distributed and defined by its mean and covariance $\boldsymbol{\Sigma}_j$; expressed in the $i$-th frame. It is usually represented by $(dx, dy, d\theta)^T$ and $(dx, dy, dz, d\omega_x, d\omega_y, d\omega_y)^T$ for the 2D and 3D cases, respectively. Also note that the hat operator maps elements from the real vector space to that of the Lie algebra $\mathfrak{se}(n)$, i.e. $\hat{\cdot}: \mathbb{R}^n\to \textbf{H}\in\mathbb{R}^{n\times n}$.

Equivalently, the estimation error can be expressed in the global frame, as it might also be done in differential representations \cite{barfoot14}:
\begin{equation}
    \mathbf{T}_{wi} =  \exp{(\hat{\mathbf{d}}_{wi})} \ \bar{\mathbf{T}}_{wi}
\end{equation}

Both uncertainty perturbation variables are related by the adjoint action, that transforms a vector from the tangent space around one element to that of another:
\begin{equation}
    \mathbf{d}_i = Ad_{\bar{\mathbf{T}}_{iw}} \ \mathbf{d}_{wi} \Leftrightarrow    \mathbf{d}_{wi} = Ad_{\bar{\mathbf{T}}_{wi}} \ \mathbf{d}_i
\end{equation}
\noindent where $Ad_{\bar{\mathbf{T}}_{wi}}$ is the adjoint representation of $\bar{\mathbf{T}}_{wi}$.

Consider now two noisy poses $\mathbf{T}_{wi}$ and $\mathbf{T}_{wk}$. Then, the relative transformation between them will be given by:
\begin{align}
    \mathbf{T}_{ik} &= \mathbf{T}_{wi}^{-1} \mathbf{T}_{wk} = \bar{\mathbf{T}}_{wi}^{-1} \ \exp{(\hat{\mathbf{d}}_{wi})}^{-1} \exp{(\hat{\mathbf{d}}_{wk})} \ \bar{\mathbf{T}}_{wk}
\end{align}

Using the following definition of the adjoint operator: $\exp{(Ad_{\bar{\mathbf{T}}_{wi}^{-1}} \ \hat{\mathbf{d}}_{wi})} \triangleq \bar{\mathbf{T}}_{wi}^{-1} \ \exp{(\hat{\mathbf{d}}_{wi})} \ \bar{\mathbf{T}}_{wi}$, and the first-order approximation \cite{brossard17} of the Baker-Campbell-Haussdorf formula for the product of exponential maps, it yields
\begin{align}
    \mathbf{T}_{ik} &\simeq \exp{(-Ad_{\bar{\mathbf{T}}_{wi}^{-1}} \ \hat{\mathbf{d}}_{wi} + Ad_{\bar{\mathbf{T}}_{wi}^{-1}} \ \hat{\mathbf{d}}_{wk})} \ \bar{\mathbf{T}}_{wi}^{-1}  \ \bar{\mathbf{T}}_{wk}\\
                    &= \exp{(-\hat{\mathbf{d}}_i + \hat{\mathbf{d}}_{ik})} \ \bar{\mathbf{T}}_{wi}^{-1}  \ \bar{\mathbf{T}}_{wk} = \exp{(\hat{\mathbf{d}})} \ \bar{\mathbf{T}}
\end{align}

The estimation error of the composed pose, in the common reference frame ($w$) and in the real vector space, will be:
\begin{align}
    \mathbf{d}^w &\simeq Ad_{\bar{\mathbf{T}}_{wi}} \  (- \mathbf{d}_i +  \mathbf{d}_{ik})\\
                    &= \begin{pmatrix} -\pmb{\mathbb{I}}_{\ell\times \ell} & \pmb{\mathbb{I}}_{\ell\times \ell} \end{pmatrix} \begin{pmatrix} \mathbf{d}_{wi} & \mathbf{d}_{wk} \end{pmatrix}^T \label{eq:1}
\end{align}
\noindent with $\mathbb{I}_{\ell\times \ell}$ the identity matrix of size $\ell$.

Generalizing \eqref{eq:1} for a full pose-graph instead of just a pair of vertices, the random vector associated to the estimation error between two arbitrary vertices $(i,k)$ which are related through the $j$-th edge can be written as:
\begin{align}
     \mathbf{d}_j \equiv \mathbf{d}_j^w \simeq \pmb{\mathbb{I}}_j \left(\mathbf{d}_{w1} \dots \mathbf{d}_{wi} \dots \mathbf{d}_{wk} \dots \mathbf{d}_{wn} \right)^T =  \pmb{\mathbb{I}}_j \boldsymbol{\delta}_w \label{eq:2}
\end{align}
\noindent where $\pmb{\mathbb{I}}_j$ is the  $1\times n$  block  matrix populated with zero matrices everywhere but in $[\pmb{\mathbb{I}}_j]_{i,k} = -[\pmb{\mathbb{I}}_j]_{k,i} = \pmb{\mathbb{I}}_{\ell\times \ell}$.

As mentioned before, once the graph is built, ML methods aim to find the set of poses that minimize the constraints introduced either by odometry or loop closures, i.e.,
\begin{equation}
    \mathbf{x}^* = \arg\min_\mathbf{x} \mathbf{F}(\mathbf{x})
\end{equation}
\noindent For that purpose, the following cost function (negative log-likelihood) is to be minimized, assuming all constraints to be independent from each other,
\begin{equation}
    \mathbf{F}(\mathbf{x}) = \sum_{j=1}^m \mathbf{F}_j(\mathbf{x}) = \sum_{j=1}^m \mathbf{d}_j^T(x) \boldsymbol{\Sigma}_j^{-1} \mathbf{d}_j(x)
\end{equation}

Revisiting \eqref{eq:2}, the cost function of an arbitrary edge may be expressed as:
\begin{align}
    \mathbf{F}_j \simeq \boldsymbol{\delta}_w^T \ \pmb{\mathbb{I}}_j^T \ \boldsymbol{\Sigma}_j^{-1} \ \pmb{\mathbb{I}}_j \ \boldsymbol{\delta}_w = \boldsymbol{\delta}_w^T \ \mathbf{Y}_j \ \boldsymbol{\delta}_w
\end{align}

Finally, the Kronecker product allows to rewrite \eqref{eq:0} in terms of the pose-graph's structure, as:
\begin{equation}
    \mathbf{Y} \simeq \sum_{j=1}^m \mathbf{E}_j \otimes \boldsymbol{\Sigma}_j^{-1} \label{eq:3}
\end{equation}

Two special cases of the above equation are worth studying, in which it is possible to directly relate $\mathbf{Y}$ and the graph Laplacian. The first one deals with the situation in which the covariance matrix is constant through measurements; a common assumption in related literature \cite{khosoussi14}, although rarely represents reality and is consistent only during purely exploratory trajectories. Under this hypothesis, \eqref{eq:3} can be particularized as:
\begin{align}
    \mathbf{Y} &\simeq \sum_{j=1}^m \mathbf{E}_j \otimes \bar{\boldsymbol{\Sigma}}^{-1} = \mathbf{L} \otimes \bar{\boldsymbol{\Sigma}}^{-1} &\text{iff } \boldsymbol{\Sigma}_{j}=\bar{\boldsymbol{\Sigma}} \ \forall j \label{eq:4}
\end{align}

The second case considers the graph's FIM to have an upper bound for every edge in the trajectory, that is,
\begin{align}
    \mathbf{Y} \simeq \sum_{j=1}^m \mathbf{E}_j \otimes \boldsymbol{\Sigma}_j^{-1} &\preceq \mathbf{Y}_{upper}
\end{align}

\noindent Implying that a lower bound in the covariance matrix of the $j$-th edge need to be considered:
\begin{equation}
    \boldsymbol{\Sigma}_j^{-1}\preceq\gamma_j\bar{\boldsymbol{\Sigma}}^{-1}  \ \ \Leftrightarrow \ \ \boldsymbol{\Sigma}_j\succeq\beta_j\bar{\boldsymbol{\Sigma}} \label{eq:bound}
\end{equation}
\noindent where $\beta_j$ and $\gamma_j$ are positive constants and $\mathbf{A}\succeq \mathbf{B}$ means $\mathbf{A}-\mathbf{B}$ is positive semidefinite. This guess would make $\beta_j$ to increase during exploratory trajectories and decrease when a loop is closed. Note that an upper bound in the covariance matrix might be useful to know when uncertainty is too big to continue exploration (stopping criteria); whilst a lower bound would be appropriate to evaluate the usefulness of loop closures (exploration-exploitation dilemma).

Under this assumption, \eqref{eq:3} becomes:
\begin{align}
    \mathbf{Y} &\preceq \sum_{j=1}^m \mathbf{E}_j \otimes \left(\gamma_j \bar{\boldsymbol{\Sigma}}^{-1} \right)&\\
    &= \mathbf{L}_w \otimes \bar{\boldsymbol{\Sigma}}^{-1} \ \ \ \text{iff } \boldsymbol{\Sigma}_j^{-1}\preceq\gamma_j\bar{\boldsymbol{\Sigma}}^{-1} \ \forall j \label{eq:5}
\end{align}
\noindent where $\mathbf{L}_w$ is the Laplacian of the graph in which each edge is weighted with $w_j=\gamma_j$. Note that \eqref{eq:4} yields as a particular case of the previous equation when $w_j=1 \ \forall j$.

One smart and mathematically-consistent way to establish that bound is via the FIM's eigenvalues, since any positive semidefinite matrix can be considered to be trivially upper-bounded by a diagonal matrix with its largest eigenvalue as diagonal terms. Let $(\rho^j_1,\dots,\rho^j_\ell)$ be the set of eigenvalues of $\boldsymbol{\Sigma}_j$, ranked in increasing order, and $(1/\rho^j_\ell,\dots,1/\rho^j_1)$ that of $\boldsymbol{\phi}_j$. Then, the bound in \eqref{eq:bound} can be expressed as:
\begin{equation}
    \boldsymbol{\Sigma}_j^{-1} \preceq \frac{1}{\rho^j_1} \ \pmb{\mathbb{I}}_{\ell\times\ell} \label{eq:bound_eopt}  \ \ \Leftrightarrow \ \ \boldsymbol{\Sigma}_j \succeq \rho^j_1 \ \pmb{\mathbb{I}}_{\ell\times\ell}
\end{equation}

\subsection{On the Spectra}

Consider now $(\bar{\rho}_1,\dots,\bar{\rho}_\ell)$ to be the set of eigenvalues of $\bar{\boldsymbol{\Sigma}}^{-1}$, and $(0=\ptwiddle{\mu}_1, \ptwiddle{\mu}_2, \dots,\ptwiddle{\mu}_n)$ that of $\mathbf{L}_{(w)}$, again ranked in increasing order. According to the spectral properties of the Kronecker product:
\begin{equation}
   \text{eig}\left(\mathbf{L}_{(w)}\otimes\bar{\boldsymbol{\Sigma}}^{-1}\right)= \ptwiddle{\mu}_k\ \bar{\rho}_b, \ \ \begin{array}{ll}
            k=1,\dots,n \\
            b=1,\dots,\ell
        \end{array}
\end{equation}

Therefore, and while uncertainty keeps constant through measurements, it can be stated that any utility function applied to $\mathbf{Y}$ can be found by applying it separately to the reduced Laplacian (after removing its zero eigenvalue) and to the edge's Information matrix:
\begin{align}
    \therefore \|\mathbf{Y}\|_p &\propto \|\mathbf{L}\|_p &\text{iff } \boldsymbol{\Sigma}_{j}=\bar{\boldsymbol{\Sigma}} \label{eq:6}
\end{align}

\noindent It can be proven as follows for $p\neq0,p\leq1$,
\begin{align}
    \|\mathbf{Y}\|_p \simeq \left(\frac{1}{n\ell} \sum\limits_{k=2}^{n} \sum\limits_{b=1}^\ell (\mu_k \bar{\rho}_b)^p \right)^\frac{1}{p} = \|\mathbf{L}\|_p \|\bar{\boldsymbol{\Sigma}}^{-1}\|_p
\end{align}
\noindent and for $p=0$,
\begin{align}
    \|\mathbf{Y}\|_p &\simeq \exp \left(\frac{1}{n\ell} \sum\limits_{k=2}^n \sum\limits_{b=1}^\ell \log(\mu_k\bar{\rho}_b) \right) = \|\mathbf{L}\|_p \|\bar{\boldsymbol{\Sigma}}^{-1}\|_p^{\frac{n-1}{n}} \label{eq:61}
\end{align}

Similarly, the following inequality is satisfied for the case all measurements are bounded as in \eqref{eq:5}:
\begin{align}
    \therefore \|\mathbf{Y}\|_p &\leq \|\mathbf{L}_w\|_p \ \textrm{with}\ w_j=\gamma_j \ \textrm{iff} \ \boldsymbol{\Sigma}_j^{-1}\preceq\gamma_j\bar{\boldsymbol{\Sigma}}^{-1} \forall j \label{eq:7}
\end{align}

Equations \eqref{eq:6} and \eqref{eq:7} allow to shift the traditional approach of computing optimality criteria on $\mathbf{Y}$, to a new strategy where they are computed on $\mathbf{L}_{(w)}$. At this point, and since these utility functions are functionals of the eigenvalues of $\mathbf{L}_{(w)}$, the Spectral Graph Theory can be leveraged. The simplest metric broadly studied in the literature is the sum its eigenvalues, that has some well-known bounds \cite{ganie19}. For the particular case of the sum of all non-zero eigenvalues, it is exactly:
\begin{align}
    S_2 (\boldsymbol{\mathcal{G}}) &\triangleq \sum\limits_{k=2}^n \mu_k = \text{trace}(\mathbf{L}) = 2m
\end{align}
\noindent Despite the interest here is in the eigenvalues of $\mathbf{L}_w$, and not in those of $\mathbf{L}$, their traces can be easily proven to be $\bar{w}$-proportional.

A second important index is the (weighted) number of spanning trees, which provides a measure for the global reliability of the graph. According to Kirchhoff's (weighted) Matrix-Tree Theorem, it is given by the determinant of the reduced Laplacian, equal to any cofactor of $\textbf{L}_{(w)}$:
\begin{equation}
    \ptwiddle{t}(\boldsymbol{\mathcal{G}}) \triangleq \text{cof}(\mathbf{L}_{(w)}) =  \frac{1}{n}\prod\limits_{k=2}^n \ptwiddle{\mu}_k
\end{equation}

The second smallest eigenvalue of the (weighted) Laplacian is also an important index of a graph, since its value reflects if the graph is disconnected \cite{de07}. It is known as the (weighted) algebraic connectivity or Fiedler value, and is greater than zero only for connected graphs:
\begin{equation}
    \ptwiddle{\alpha}(\boldsymbol{\mathcal{G}}) \triangleq \min(\ptwiddle{\mu}_k : k=2,\dots,n)=\ptwiddle{\mu}_2
\end{equation}

Lastly, the Kirchhoff index measures the resistance between each pair of vertices assuming edges as unit resistors, and is defined by: $Kf(\boldsymbol{\mathcal{G}}) \triangleq n\sum\limits_{k=2}^n \mu_k^{-1}$.

Therefore, for the most general case in which covariance is bounded, uncertainty's optimality criteria can be expressed in terms of the graph's structure as:
\begin{align}
    &T{\text -} opt(\mathbf{Y}) &= A{\text -} opt(\boldsymbol{\Sigma})^{-1} &\leq T{\text -} opt (\mathbf{L}_w) = \frac{2m}{n}\ \bar{w}\\
    &D{\text -} opt(\mathbf{Y}) &= D{\text -} opt(\boldsymbol{\Sigma})^{-1} &\leq D{\text -} opt (\mathbf{L}_w) = \left( n\ \tilde{t}(\boldsymbol{\mathcal{G}})\right)^{\frac{1}{n}}\label{eq:dopt_graph}\\
    &A{\text -} opt(\mathbf{Y}) &= T{\text -} opt(\boldsymbol{\Sigma})^{-1} &\leq A{\text -} opt (\mathbf{L}_w) = (*) \label{eq:uncomplete}\\
    &E{\text -} opt(\mathbf{Y}) &= \tilde{E}{\text -} opt(\boldsymbol{\Sigma})^{-1} &\leq E{\text -} opt (\mathbf{L}_w) = \tilde{\alpha}(\boldsymbol{\mathcal{G}})
\end{align}
\noindent Note that \eqref{eq:uncomplete}, dependent of the weighted Kirchhoff index, is not complete. Although it indeed is known to have some bounds \cite{zhu13}, their computation cost makes its use worthless and thus it has been omitted. For the particular case in which $\boldsymbol{\Sigma}_{j}=\bar{\boldsymbol{\Sigma}} \ \forall j$, $(*)=n^2 / Kf(\boldsymbol{\mathcal{G}})$. Also note that results are consistent with those reported by \cite{khosoussi14} for the particular case they studied where $\ell=\{2, 3\}$, covariance is constant through measurements and $D{\text -} opt$ is defined in a traditional way \cite{wald43}.

\section{Offline Experimental Results}\label{S:4}

In this section, several experiments are shown in order to prove the relationships proposed hereinabove. Datasets from \cite{carlone14} and \cite{carlone15} have been used for 2D and 3D experiments, respectively.

\subsection{Constant Uncertainty along Edges}

A first experiment has been carried out to prove the relationship between $\|\mathbf{Y}\|_p$ and $\|\mathbf{L}\|_p$ when uncertainty keeps constant along the trajectory. Suppose the std. deviation of the linear movement to be $\sigma_{x,y}=0.3$ m, and $\sigma_\theta=0.063$ rad that of the orientation. Then, the Information matrix of any edge will be  $\bar{\boldsymbol{\phi}} = \text{diag}(11.11, 11.11, 250)$.

\begin{figure}[t!]
      \vspace{2mm}
      \centering
      \includegraphics[max height=3cm,max width=\linewidth]{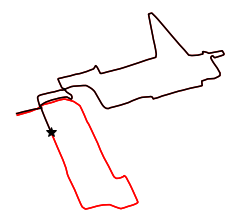}
      \caption{Complete (black) and reduced (red) trajectories of the FRH dataset. A star denotes the starting position.}
      \label{fig:FRH_P}
\end{figure}

To compare both methods to compute optimality criteria, the previous FIM has been assigned to every odometry edge in FRH dataset. However, instead of the whole dataset, a reduced exploratory trajectory has been used, which contains the first 400 nodes. Fig. \ref{fig:FRH_P} shows the complete and reduced trajectories of this sequence, consisting of 1316 nodes and 1485 constraints.

Figure \ref{fig:FRH_comparison_fim} shows $T\text{-},D\text{-}$ and $E\text{-}opt$ criteria computed by both approaches. Blue curves correspond to applying optimality criteria to $\mathbf{Y}$, while red ones do to applying them to the Laplacian and $\bar{\boldsymbol{\phi}}$ matrices. Note that all curves overlap, proving that equalities in \eqref{eq:6}-\eqref{eq:61} hold. It is also worth noticing that, under this assumption, it is not possible to capture the behavior during loop closures.

\subsection{Variable Uncertainty along Edges}

Consider now the second and more realistic case in which $\boldsymbol{\phi}_j$ is not constant anymore. For that purpose, a weighted graph has to be built, where the weight of the $j$-th edge will be $w_j=\| \boldsymbol{\phi}_j\|_{\infty}\equiv\tilde{E}\text{-}opt(\boldsymbol{\phi}_j)$, as expressed in \eqref{eq:bound_eopt}. In this case, the whole trajectory contained in the FRH dataset has been studied, inasmuch as it is tractable despite being computationally intensive.

\begin{figure*}
  \vspace{2mm}
  \begin{subfigure}[t]{0.3\linewidth}
      \centering
      \includegraphics[max height=4cm,max width=\linewidth]{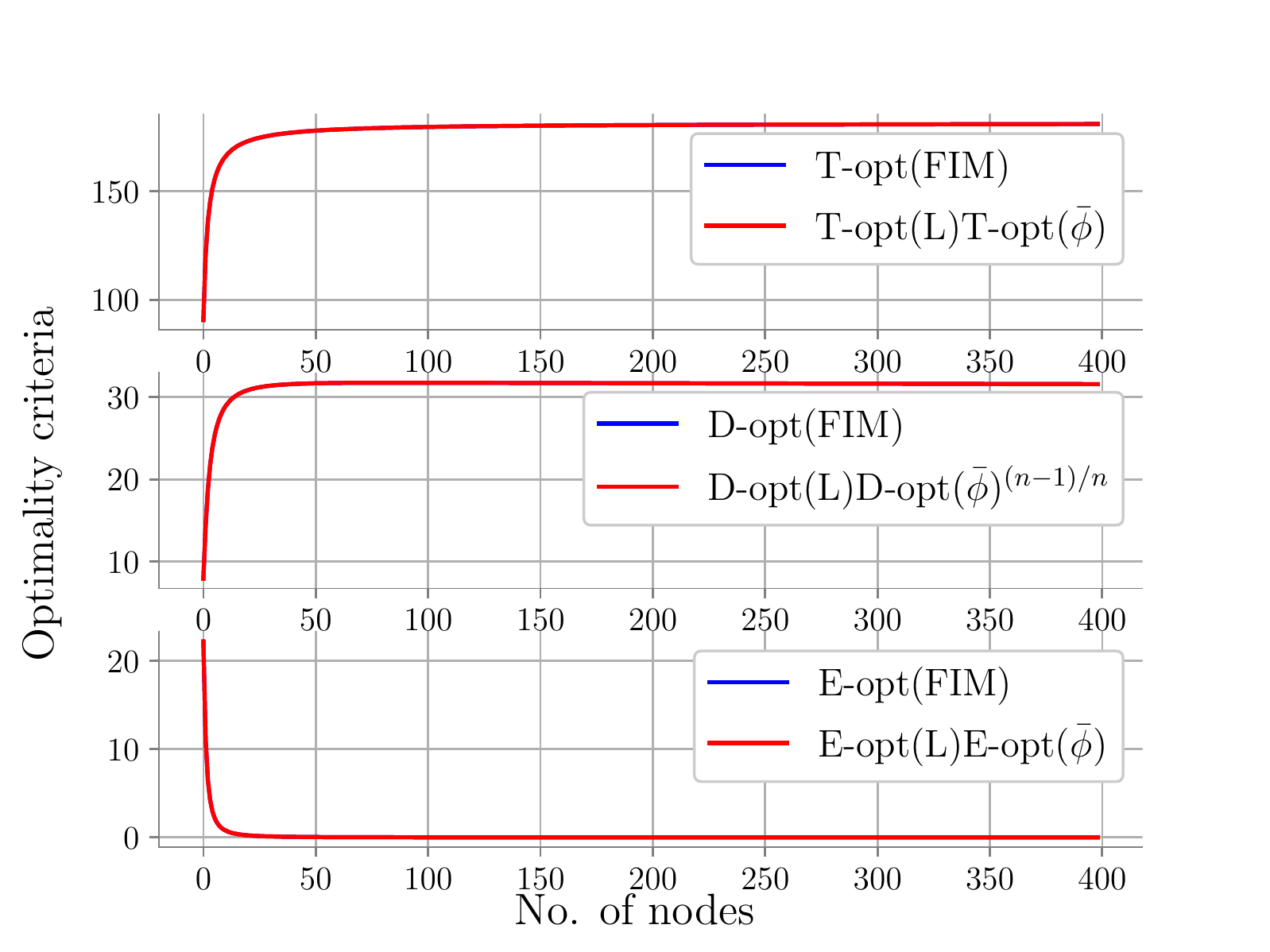}
      \caption{}
      \label{fig:FRH_comparison_fim}
  \end{subfigure} \hfill
  \begin{subfigure}[t]{0.3\linewidth}
      \centering
      \includegraphics[max height=7cm,max width=\linewidth]{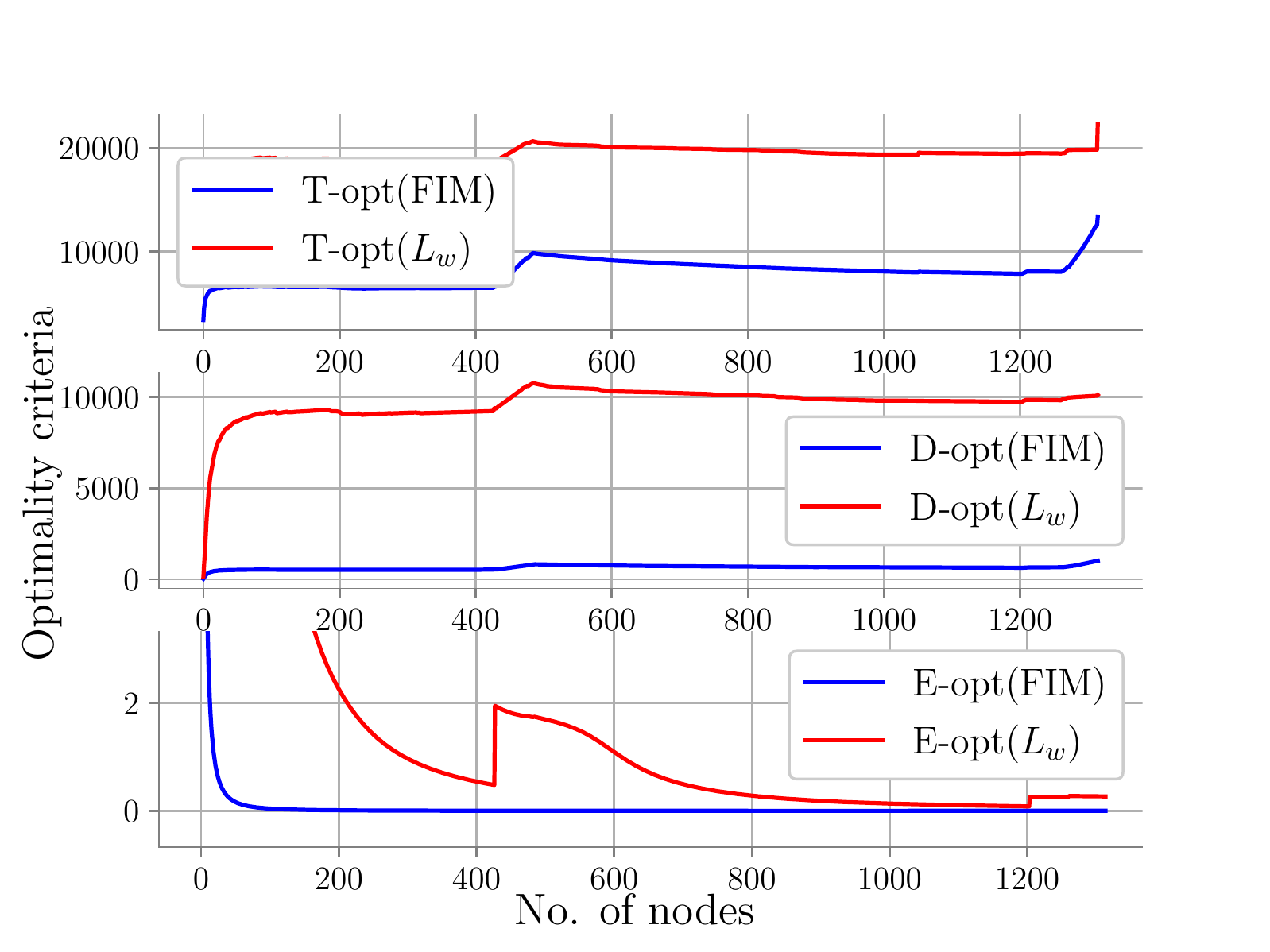}
      \caption{}
      \label{fig:FRH_comparison_FIM_var_full}
  \end{subfigure} \hfill
  \begin{subfigure}[t]{0.3\linewidth}
      \centering
      \includegraphics[max height=7cm,max width=\linewidth]{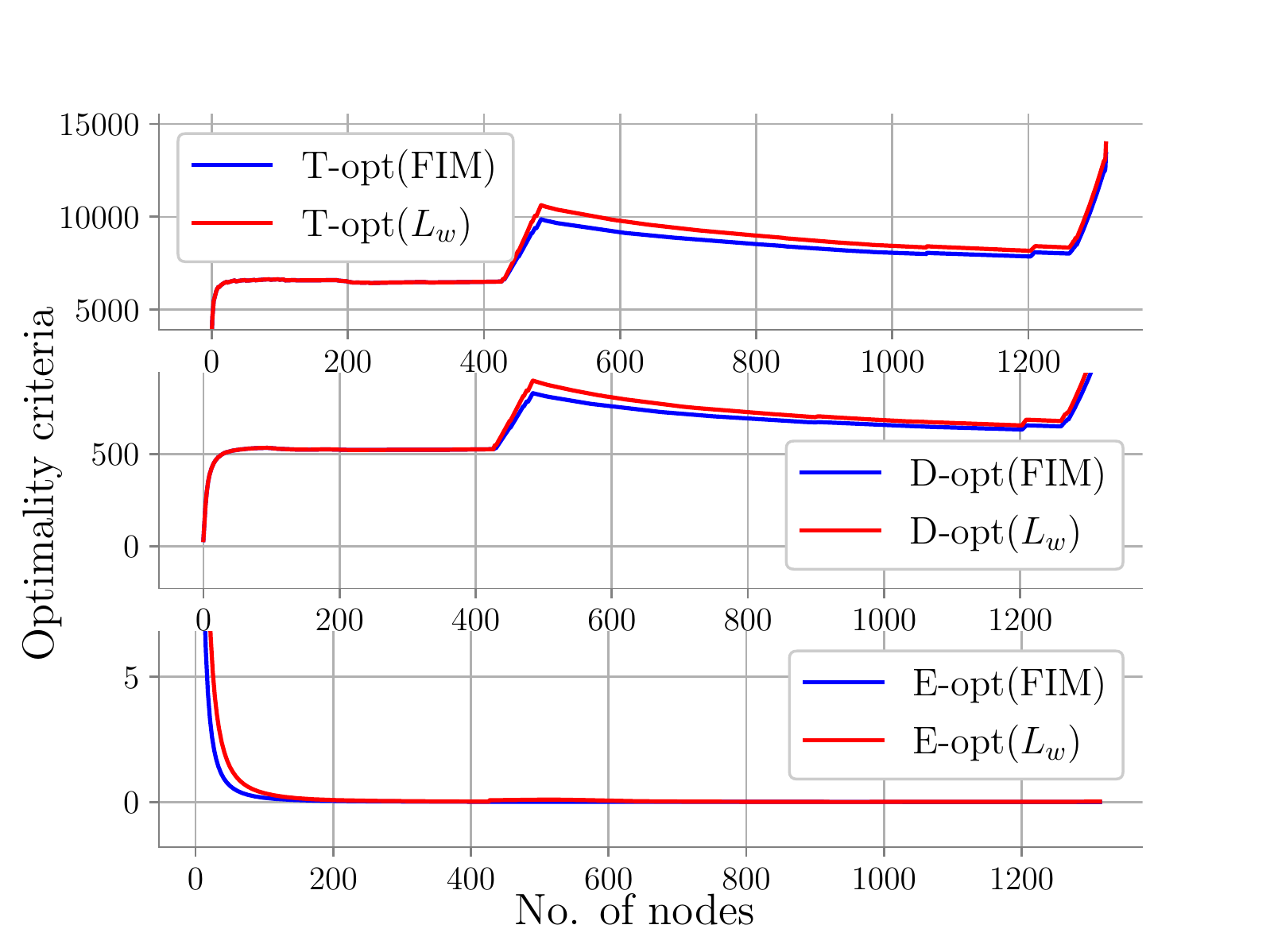}
      \caption{}
      \label{fig:MITb_comparison_FIM_var_full_newbounds}
  \end{subfigure} \hfill
  \caption{Results of optimality criteria computation on the FRH dataset, based on the FIM (blue) and the Laplacian (red). (a) contains results for the reduced trajectory in which uncertainty is constant; while (b) and (c) correspond with variable uncertainty, using $w_j=\|\boldsymbol{\phi}_j\|_\infty$ and $w_j=\|\boldsymbol{\phi}_j\|_p$, respectively.}
\end{figure*}

Figure \ref{fig:FRH_comparison_FIM_var_full} contains the resulting $T\text{-},D\text{-}$ and $E\text{-}opt$ for both $\mathbf{Y}$ and $\mathbf{L}_w$. Effectively, red curves act as upper limit of the blue ones, due to the chosen weights. That bound indeed limits $\|\mathbf{Y}\|_p \ \forall p$, though it is an extremely conservative one. The chosen limit made $\tilde{E}\text{-}opt(\mathbf{Y})$ and $\tilde{E}\text{-}opt(\mathbf{L}_w)$ to be approximately equal. And since $\tilde{E}\text{-}opt\geq T\text{-}opt\geq D\text{-}opt\geq A\text{-}opt\geq E\text{-}opt$, it represents an upper limit for every other criteria. The choice of the Laplacian's weights are therefore a key issue in the graph representation, and could be set in order to be more akin to certain optimality criteria. Equation \eqref{eq:bound_eopt} may be rewritten as follows so as to approximate optimality criteria more accurately: $\boldsymbol{\phi}_j \sim \|\boldsymbol{\phi}_j\|_p \ \pmb{\mathbb{I}}_{\ell\times\ell}$. In this case, the following relationship will hold,
\begin{equation}
    \| \mathbf{Y}\|_p \sim \|\mathbf{L}_w\|_p \ \ \textrm{with} \ \  w_j=\|\boldsymbol{\phi}_j\|_p
\end{equation}

Figure \ref{fig:MITb_comparison_FIM_var_full_newbounds} shows the results using the new weights. It is easy to notice how curves are now much closer to each other. Interestingly, they overlap in the exploratory trajectory that occurs during the first 420 steps since the same FIM was associated to those edges in the dataset (constant uncertainty case). Small deviations observed in $E\text{-}opt$ are attributed to numerical issues with minimum eigenvalue computations.

\begin{figure*}
  \begin{subfigure}[t]{0.3\linewidth}
      \centering
      \includegraphics[max height=4cm,max width=\linewidth]{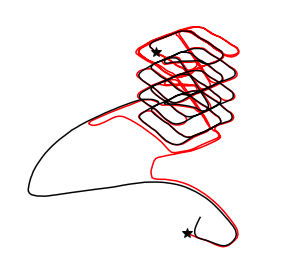}
      \caption{}
      \label{fig:garage}
  \end{subfigure} \hfill
  \begin{subfigure}[t]{0.3\linewidth}
      \centering
      \includegraphics[max height=7cm,max width=\linewidth]{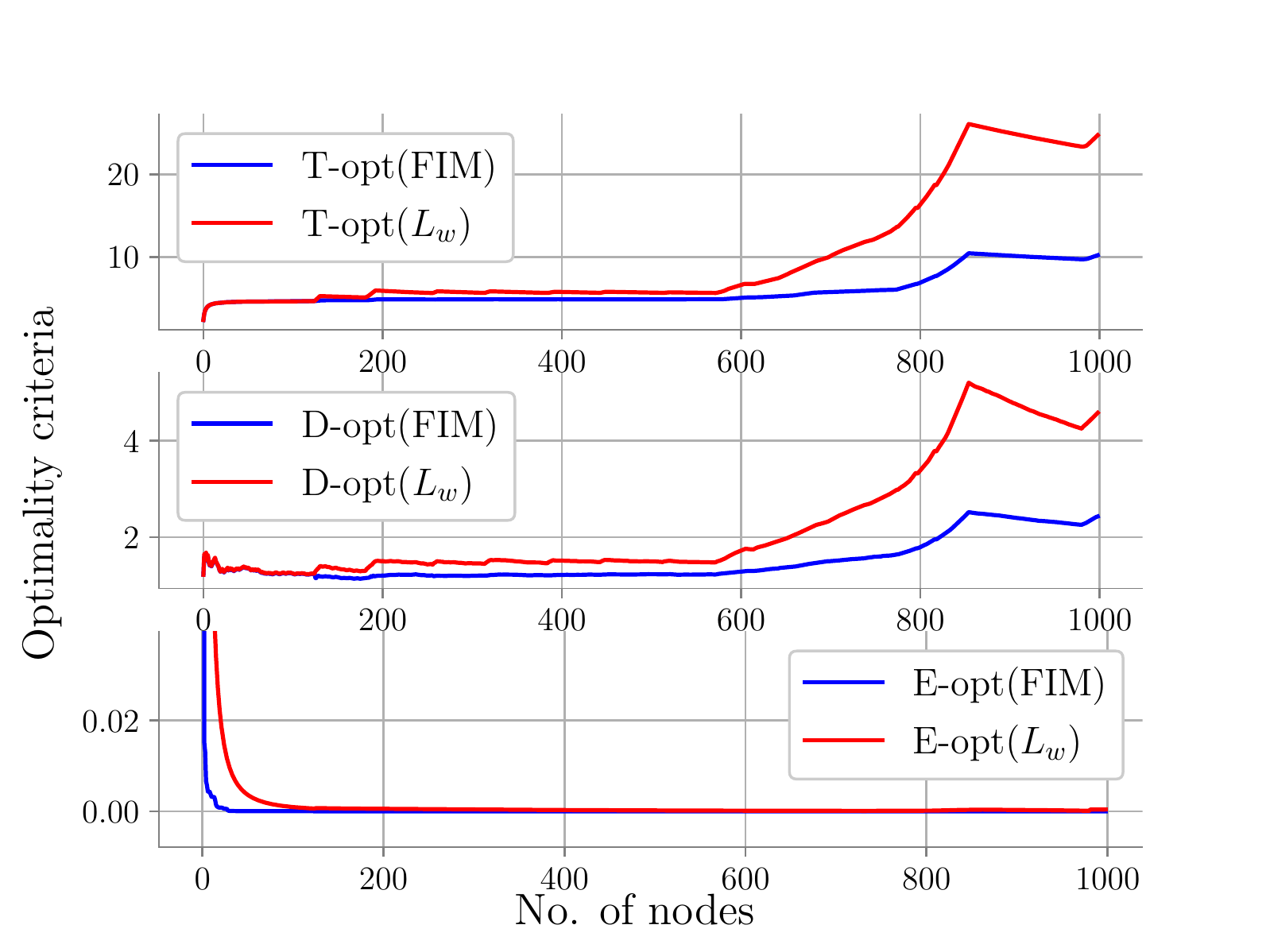}
      \caption{}
      \label{fig:garage_comparison}
  \end{subfigure} \hfill
  \begin{subfigure}[t]{0.3\linewidth}
      \centering
      \includegraphics[max height=7cm,max width=\linewidth]{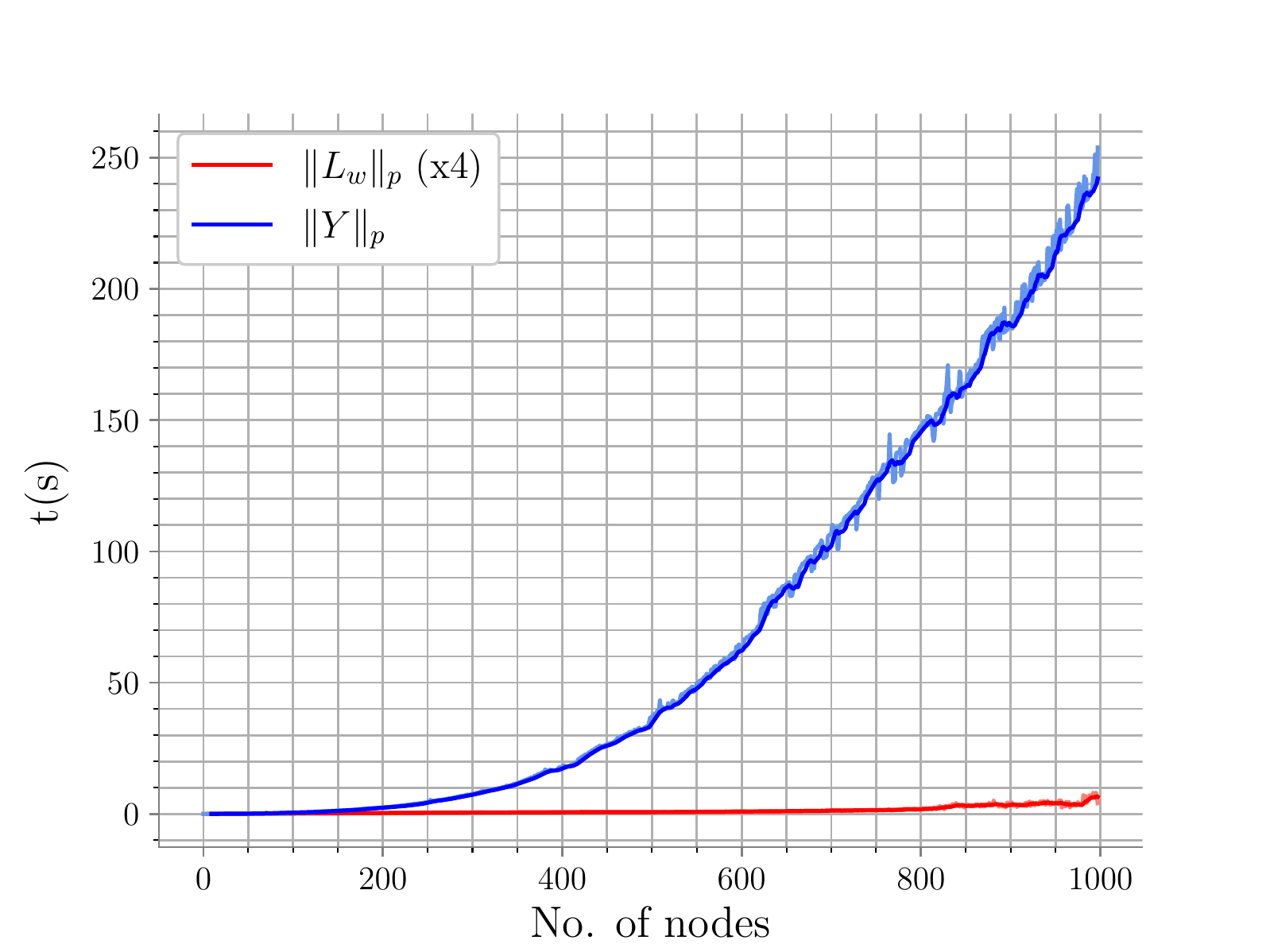}
      \caption{}
      \label{fig:garage_time}
  \end{subfigure} \hfill
  \caption{(a) Complete (black) and reduced (red) trajectories of the Garage 3D dataset, (b) results of optimality criteria computation, based on the FIM (blue) and the weighted Laplacian with $w_j=\|\boldsymbol{\phi}_j\|_p$ (red), and (c) time consumed at each step by both approaches.}
\end{figure*}

Finally, an experiment has been performed in the 3D Garage dataset, to prove that the proposed relationships hold and are independent of $\ell$. This dataset contains 1661 nodes and 6275 constraints, and its trajectory is depicted in Fig. \ref{fig:garage}. Due to the computational complexity, only the first 1000 nodes (over 2600 edges) have been used in this experiment (red trajectory in that figure). Results of the computation of $T\text{-},D\text{-}$ and $E\text{-}opt$ by both approaches appear in Fig. \ref{fig:garage_comparison}. $\| \mathbf{L}_w\|_p$ and $\| \mathbf{Y}\|_p$ are approximately equal during most part of the processed trajectory, and when they differ, $\| \mathbf{L}_w\|_p$ is maintained as an upper bound; an interesting behavior which reinforces its use during Active SLAM. Figure \ref{fig:garage_time} shows the evolution of the time consumed to compute utility functions as the pose-graph grows. Computations over $\mathbf{Y}$ led to significantly higher time, taking up to 4 minutes per step to analyze the full FIM at the end of the sequence's patch. Note that red curves are scaled by four. Studying the computational complexity of both methods, it is easy to notice that computing optimality criteria on $\mathbf{L}_w$ is $\mathcal{O}(n^3)+\mathcal{O}(\ell^3)$ while applying them to $\mathbf{Y}$ is $\mathcal{O}(\ell^3n^3)$, omitting lower order terms and being $\mathcal{O}(\cdot)$ a lower bound.

\section{Online D-optimal Active SLAM} \label{S:5}

Previous sections have proved, both theoretically and experimentally, that analyzing the underlying weighted pose-graph is equivalent to the traditional analysis of $\| \mathbf{Y}\|_p$ in order to perform Active SLAM; with the difference of significantly lower computing time. In this section, we go a step further in this direction, performing D-optimal Active SLAM based on the analysis of the underlying pose-graph; in a complex simulation environment within a Gazebo-ROS framework (Fig. \ref{fig:active_system}).

The proposed approach contains two main modules. The first one deals with the estimation of the robot's pose and the construction of a map of the environment from laser measurements, using a slightly modified version of Open Karto \cite{konolige10} as SLAM front-end (graph construction) and g2o \cite{grisetti11} as back-end (graph optimization). Active SLAM module, which stems from \cite{umari17}, handles (i) the identification of possible locations to explore, (ii) the computation of $D\text{-}opt$ associated to the actions that would take the robot to each of those locations, and (iii) the selection and execution of the optimal set of actions. Each of these steps are carried out in different submodules, see Fig. \ref{fig:active_system}.

\subsection{Frontier Detection}
First of all, candidate locations are recognized using two frontier detectors based on Rapidly-exploring Random Trees (RRT), acting on the local and global maps, and a third one based on the Canny edge detection algorithm. All frontiers are fused and filtered, similarly as done in \cite{umari17}, but in a more efficient and restrictive fashion.

\subsection{Decision Making and Action Execution}
In order to select the most informative frontier, the following sequence is repeated for each candidate location. Firstly, the path to reach it is computed using Dijkstra's algorithm (global planner). Then, a new weighted pose-graph associated to that hallucinated path is built, adding to the last known pose-graph from the SLAM algorithm nodes distributed along the trajectory (the longer the path the greater the number of nodes) and one final node at the frontier's location. These hallucinated nodes will be connected sequentially through odometry constraints.

To our understanding, the expected FIM associated to each hallucinated edge, has to account for three factors: (i) the uncertainty increase due to the robot's movement, (ii) the information gained when exploring unknown areas and (iii) the uncertainty reduction due to potential loop closures. The following heuristic rules, akin to those used in \cite{carrillo18}, allow to predict the uncertainty along the hallucinated pose-graphs. Firstly, the last known FIM is sequentially reduced as the pose-graph grows. Secondly, the percentages of unknown and occupied areas in the neighborhood of each hallucinated node is computed. Greater percentages will encode, respectively, larger new areas to be potentially explored and a higher chance of recognizing a previously-seen textured area. Thus, the FIM is scaled again with values that encode those percentages. Then, the weights of the hallucinated pose-graph are defined as $w_j = \|\boldsymbol{\phi_j}\|_0 \equiv D\text{-}opt(\boldsymbol{\phi_j})$.

Lastly, $D\text{-}opt(\mathbf{L}_w)$ is computed for every candidate's pose-graph, according to \eqref{eq:dopt_graph}. The next goal destination will be the candidate with the greatest utility. ROS package \texttt{move\_base} deals with the planning of the path towards that frontier (global) and its execution (local planner).

\subsection{Simulation Results}
A wheeled robot equipped with a laser sensor with 180\degree \ field of view and 6 meters range has been deployed into a Gazebo's office environment, similar to Willow Garage, which encloses an area of about $56m\times45 m$.

\begin{figure*}
    \vspace{2mm}
    \centering
    \begin{subfigure}[b]{0.48\linewidth}
        \centering
        \includegraphics[max height=5cm,max width=\linewidth]{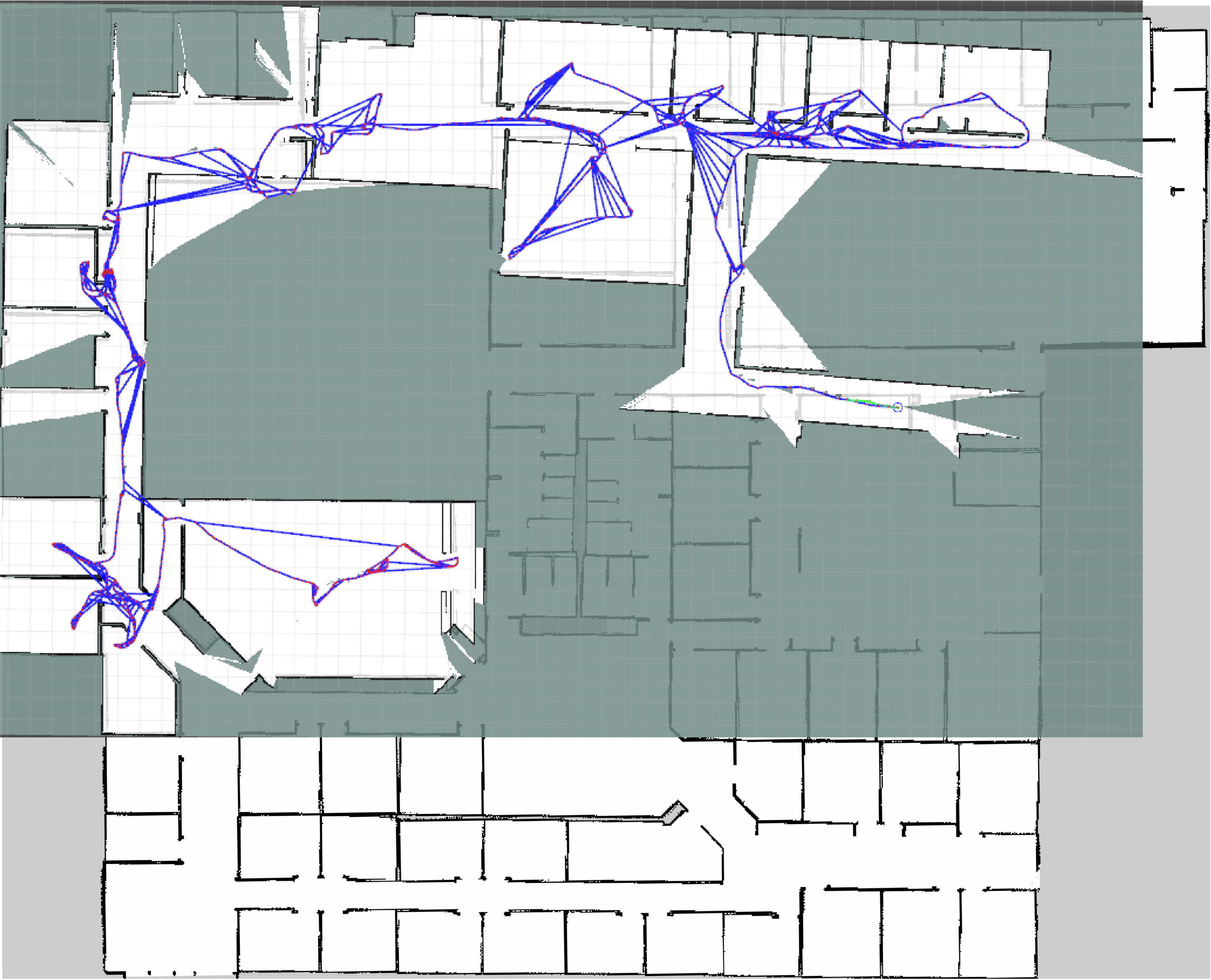}
        \caption{Closest frontier}
        \label{fig:map_closest}
    \end{subfigure} \hfill
    \begin{subfigure}[b]{0.48\linewidth}
        \centering
        \includegraphics[max height=5cm,max width=\linewidth]{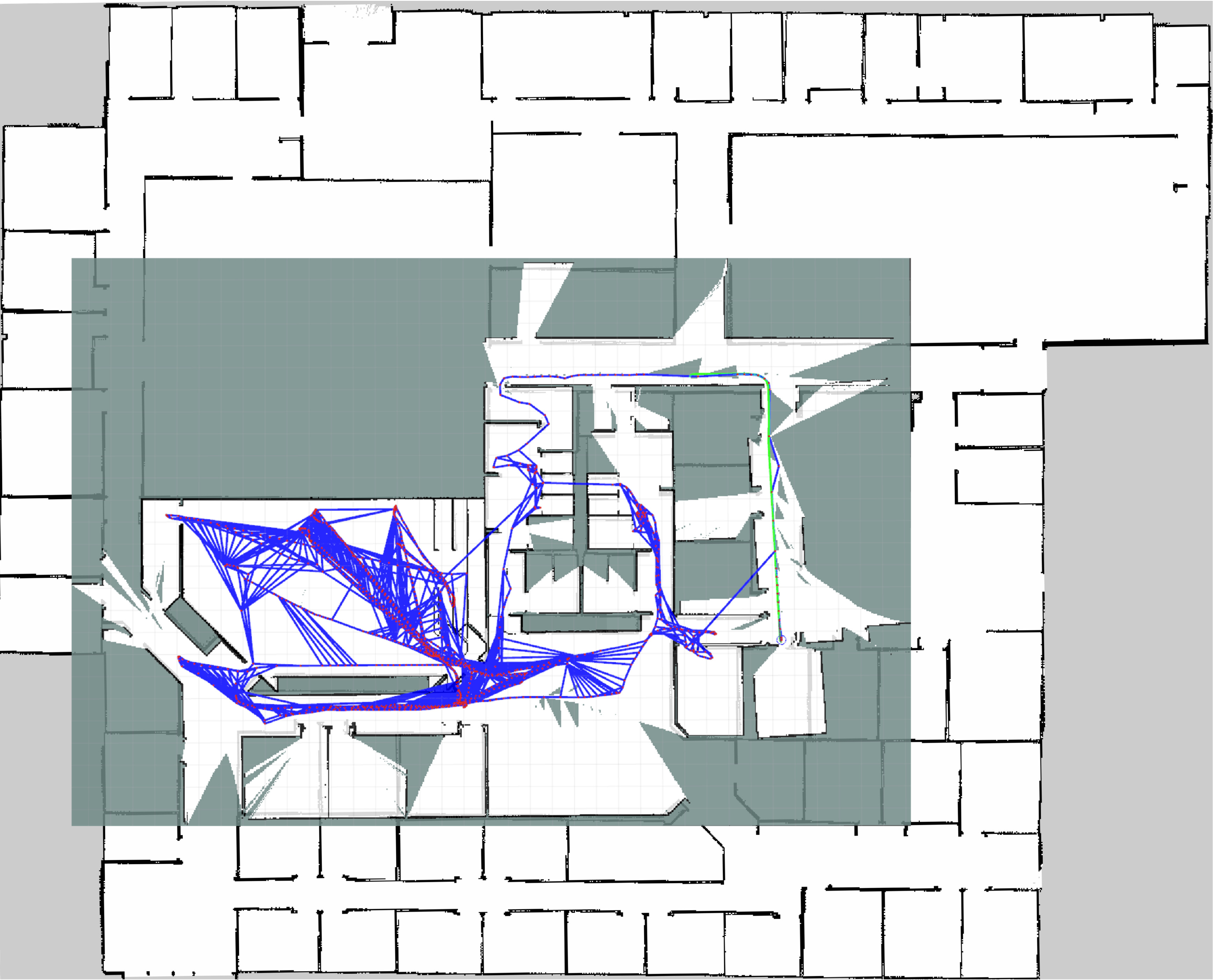}
        \caption{RRT exploration \cite{umari17}}
        \label{fig:map_rrt}
    \end{subfigure}
    \begin{subfigure}[b]{0.48\linewidth}
        \centering
        \includegraphics[max height=5cm,max width=\linewidth]{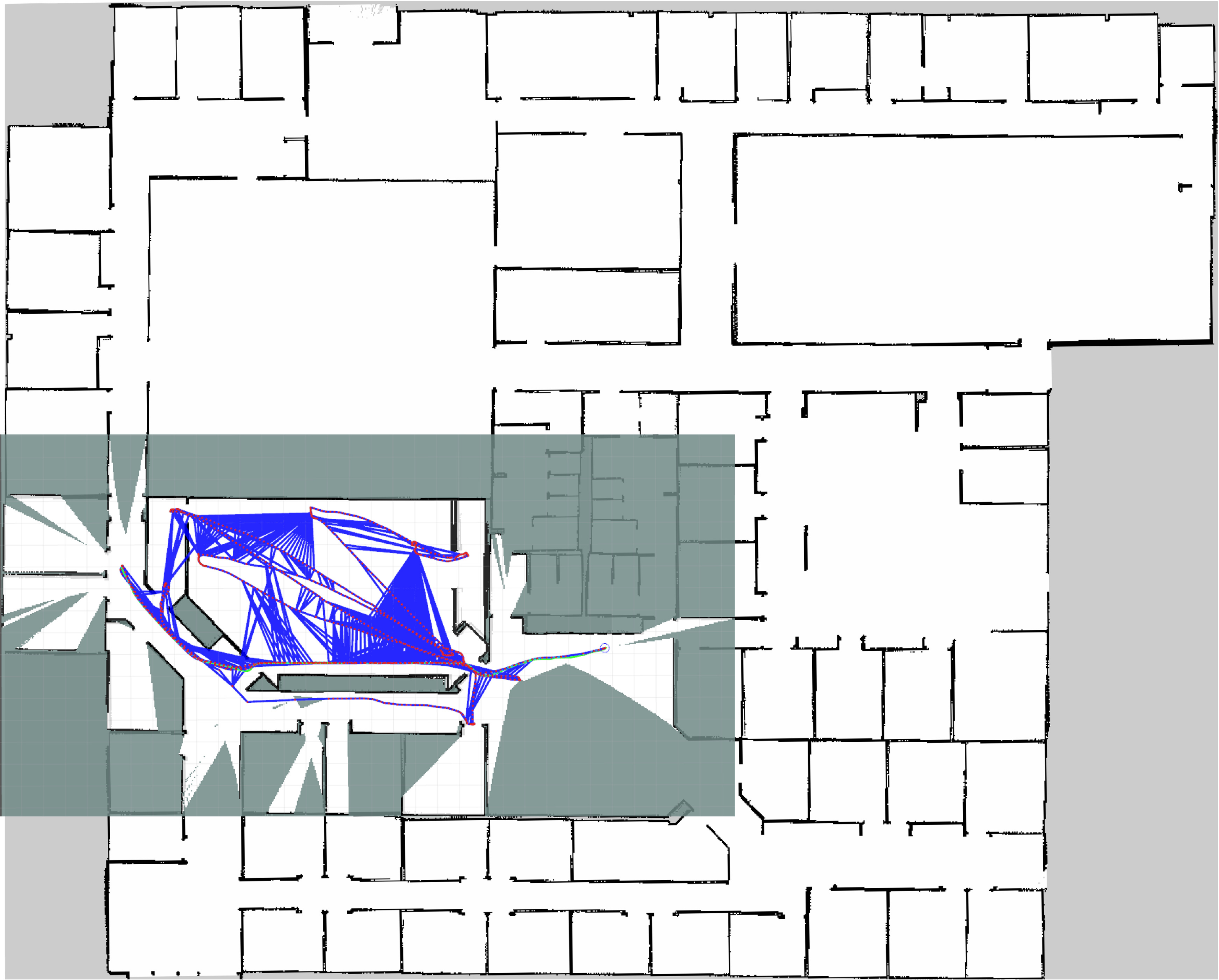}
        \caption{Shannon-R\'enyi entropy \cite{carrillo18}}
        \label{fig:map_carrillo}
    \end{subfigure} \hfill
    \begin{subfigure}[b]{0.48\linewidth}
        \centering
        \includegraphics[max height=5cm,max width=\linewidth]{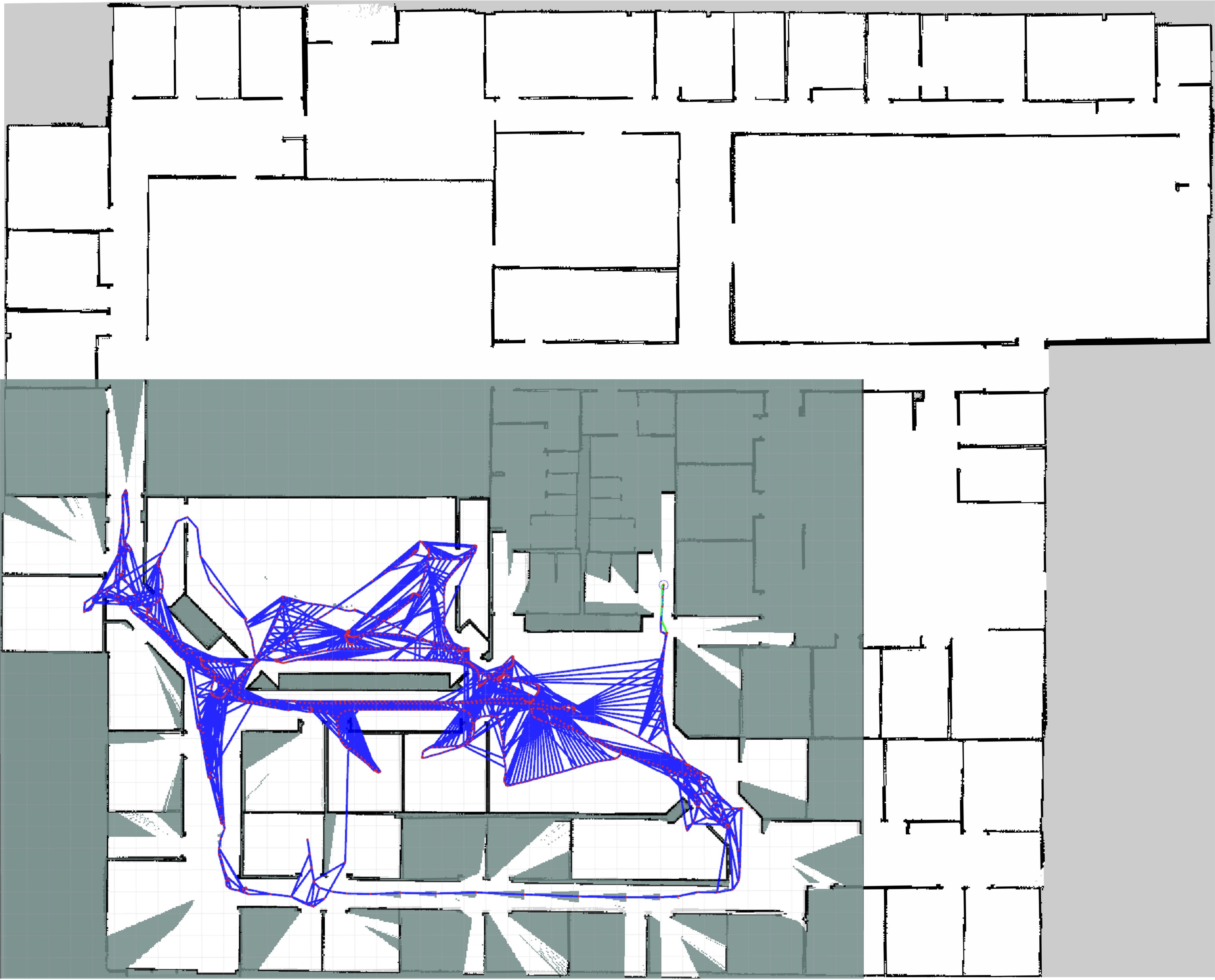}
        \caption{Graph D-opt (ours)}
        \label{fig:map_graph}
    \end{subfigure}
    \caption{Comparison of the maps and pose-graphs generated by each agent after 30 minutes of autonomous exploration. Nodes are depicted as red dots and blue lines represent constraints between them. }\label{fig:map_comparison_agents}
\end{figure*}

Figure \ref{fig:map_comparison_agents} contains the generated maps and pose-graphs after 30 minutes of autonomous exploration, plotted on top of the complete map of the environment for the ease of comparison. Four different agents have been used to benchmark our approach (IV), which share SLAM, frontier detection and navigation modules. The baseline agent (I) selects the closest frontier. Decision making of the second agent (II) corresponds to \cite{umari17}, while the third one (III) is based on the computation of Shannon-R\'enyi entropy \cite{carrillo18}. Table \ref{tab:comparison_agents} contains a quantitative comparison between agents that support qualitative results of Fig. \ref{fig:map_comparison_agents}; in terms of the size of the map and the percentage of environment explored (C), the maximum RMSE found in the map ($\epsilon$), and some metrics of the underlying graph, namely its average degree ($\bar{d}=2m/n$) and its normalized tree connectivity, $\bar{\boldsymbol{\tau}}(\boldsymbol{\mathcal{G}})$ \cite{khosoussi19}.

The baseline (I) generated a large map and explored one third of the whole environment, since it selected one path and just explored it forward. Nevertheless, the lack of exploitation made the uncertainty to grow constantly, and resulted in wrong estimates, drift and a weakly connected graph. In fact, results show that vertices are connected with 2 to 3 other vertices (corresponding 2 to purely odometry constraints). (II), the simplest agent in which actions are picked after reasoning, performed notably better. It built a more consistent map with less than half the previous $\epsilon$, although the explored area decreased due to the time spent in decision making and revisiting previously seen areas. Most nodes are clustered in the center of the map since its exploration strategy is based on going to a distant frontier and quickly returning. (III) built the most precise and consistent map, with an error  about 6 times lower than (II). However, that was achieved at the cost of exploring a very reduced area due to the time required to compute Shannon's map entropy, optimality criteria of the \textit{action pose-graph} and, finally, R\'enyi's entropy. This agent created an strongly connected pose-graph, although results need to be carefully taken into consideration due to its reduced trajectory. The proposed agent (IV) built jointly a large consistent map and a strongly connected graph. An exploration-exploitation strategy is clearly seen, similar to that of (III), but resulting in an explored area twice in size due to the relaxation of the computational load that graph evaluation allowed. Interestingly, despite the map's size is similar to the one of (II), coverage increase is above $20\%$ and map's error reduction is near $4$ times. Moreover, it increased the graph's average degree and outperformed any other agent in terms of $\bar{\tau}(\boldsymbol{\mathcal{G}})$ (the optimization target). Note that this is a normalized metric, and thus little increments on it may correspond to a high number of spanning trees depending on $n$. Our agent reached a balance between building a precise map and a strongly connected graph, and exploring a wider area; combining therefore the benefits from previous agents.

\renewcommand\thesubfigure{\Roman{subfigure}}

\begin{table}
    \vspace{2mm}
    \centering
    \begin{tabular}{l|c|c|c||c|c}
        \textbf{Agent} &\textbf{Size (m)} &\textbf{C (\%)} &$\boldsymbol{\epsilon}$ \textbf{(m)} &$\bar{\mathbf{d}}$ &$\bar{\boldsymbol{\tau}}(\boldsymbol{\mathcal{G}})$\\ [0.5ex]
        \hline
        (I) Closest             &$52\times33$           &30.38          &1.41           &2.70             &0.0744           \\ \hline
        (II) \cite{umari17}     &$38\times26$           &21.34          &0.63           &3.39             &0.1121           \\
        (III) \cite{carrillo18} &$34\times18$           &11.95          &$\mathbf{0.11}$  &$\mathbf{3.71}$    &0.1184           \\
        (IV) Ours               &$\mathbf{40\times27}$  &$\mathbf{25.79}$ &0.17           &3.49             &$\mathbf{0.1185}$  \\
    \end{tabular}
    \caption{Exploration metrics after 30 min.}
    \label{tab:comparison_agents}
\end{table}

\section{Conclusions}\label{S:6}

In this paper, we have defined the relationship between optimality criteria of the FIM and those of the Laplacian of the underlying weighted pose-graph; for the general case of Active graph SLAM formulated over Lie groups in which uncertainty evolves as the trajectory does. Therefore, the selection of optimal actions to explore an unknown environment may be achieved by exploiting the graphical facet of the problem instead of using the estimation one. We have validated the proposed relationships in 2 and 3D datasets, showing that using the Laplacian spectrum leads to equivalent behaviors and notably relaxes the computational load, even when uncertainty is not constant along the trajectory. In addition, we have presented a novel Active SLAM approach that leverages this theoretical analysis to pick D-optimal actions, leading to equivalent behaviors to other state-of-the-art methods based on expensive FIM computations in a remarkably lower time. Hence, when considering finite time exploration horizons, our agent achieves superior results.

As future work, we aim to carry out more thorough experiments and study the use of visual graph SLAM. We are also interested in using these metrics for path planning in the belief space.


\section*{Acknowledgments}
This work was supported by the Spanish government under grant PID2019‐108398GB‐I00 and by Arag\'on government under grant T45-20R.

\bibliographystyle{IEEEtran}
\bibliography{references}

\begin{thebibliography}{10}
\providecommand{\url}[1]{#1}
\csname url@samestyle\endcsname
\providecommand{\newblock}{\relax}
\providecommand{\bibinfo}[2]{#2}
\providecommand{\BIBentrySTDinterwordspacing}{\spaceskip=0pt\relax}
\providecommand{\BIBentryALTinterwordstretchfactor}{4}
\providecommand{\BIBentryALTinterwordspacing}{\spaceskip=\fontdimen2\font plus
\BIBentryALTinterwordstretchfactor\fontdimen3\font minus
  \fontdimen4\font\relax}
\providecommand{\BIBforeignlanguage}[2]{{%
\expandafter\ifx\csname l@#1\endcsname\relax
\typeout{** WARNING: IEEEtran.bst: No hyphenation pattern has been}%
\typeout{** loaded for the language `#1'. Using the pattern for}%
\typeout{** the default language instead.}%
\else
\language=\csname l@#1\endcsname
\fi
#2}}
\providecommand{\BIBdecl}{\relax}
\BIBdecl

\bibitem{thrun02}
S.~Thrun, ``Probabilistic robotics,'' \emph{Communications of the ACM},
  vol.~45, no.~3, pp. 52--57, 2002.

\bibitem{durrant06}
H.~Durrant-Whyte and T.~Bailey, ``Simultaneous localization and mapping: part
  i,'' \emph{IEEE Robotics \& Automation Magazine}, vol.~13, no.~2, pp.
  99--110, 2006.

\bibitem{grisetti10}
G.~Grisetti, R.~Kummerle, C.~Stachniss, and W.~Burgard, ``A tutorial on
  graph-based slam,'' \emph{IEEE Intelligent Transportation Systems Magazine},
  vol.~2, no.~4, pp. 31--43, 2010.

\bibitem{cadena16}
C.~Cadena, L.~Carlone, H.~Carrillo, Y.~Latif, D.~Scaramuzza, J.~Neira, I.~Reid,
  and J.~J. Leonard, ``Past, present, and future of simultaneous localization
  and mapping: Toward the robust-perception age,'' \emph{IEEE Transactions on
  Robotics}, vol.~32, no.~6, pp. 1309--1332, 2016.

\bibitem{burgard97}
W.~Burgard, D.~Fox, and S.~Thrun, ``Active mobile robot localization,'' in
  \emph{1997 International Joint Conferences on Artificial Intelligence},
  Nagoya, Japan, 1997, pp. 1346--1352.

\bibitem{feder99}
H.~J.~S. Feder, J.~J. Leonard, and C.~M. Smith, ``Adaptive mobile robot
  navigation and mapping,'' \emph{The International Journal of Robotics
  Research}, vol.~18, no.~7, pp. 650--668, 1999.

\bibitem{carrillo12}
H.~Carrillo, I.~Reid, and J.~A. Castellanos, ``On the comparison of uncertainty
  criteria for active slam,'' in \emph{2012 IEEE International Conference on
  Robotics and Automation (ICRA)}.\hskip 1em plus 0.5em minus 0.4em\relax IEEE,
  Minnesota, USA, 2012, pp. 2080--2087.

\bibitem{makarenko02}
A.~A. Makarenko, S.~B. Williams, F.~Bourgault, and H.~F. Durrant-Whyte, ``An
  experiment in integrated exploration,'' in \emph{2002 IEEE/RSJ International
  Conference on Intelligent Robots and Systems (IROS)}, vol.~1.\hskip 1em plus
  0.5em minus 0.4em\relax IEEE, 2002, pp. 534--539.

\bibitem{pukelsheim06}
F.~Pukelsheim, \emph{Optimal design of experiments}.\hskip 1em plus 0.5em minus
  0.4em\relax SIAM, 2006.

\bibitem{leung06}
C.~Leung, S.~Huang, and G.~Dissanayake, ``Active slam using model predictive
  control and attractor based exploration,'' in \emph{2006 IEEE/RSJ
  International Conference on Intelligent Robots and Systems (IROS)}.\hskip 1em
  plus 0.5em minus 0.4em\relax IEEE, 2006, pp. 5026--5031.

\bibitem{carlone14bis}
L.~Carlone, J.~Du, M.~K. Ng, B.~Bona, and M.~Indri, ``Active slam and
  exploration with particle filters using kullback-leibler divergence,''
  \emph{Journal of Intelligent \& Robotic Systems}, vol.~75, no.~2, pp.
  291--311, 2014.

\bibitem{bai16}
S.~Bai, J.~Wang, F.~Chen, and B.~Englot, ``Information-theoretic exploration
  with bayesian optimization,'' in \emph{2016 IEEE/RSJ International Conference
  on Intelligent Robots and Systems (IROS)}.\hskip 1em plus 0.5em minus
  0.4em\relax IEEE, 2016, pp. 1816--1822.

\bibitem{carrillo18}
H.~Carrillo, P.~Dames, V.~Kumar, and J.~A. Castellanos, ``Autonomous robotic
  exploration using a utility function based on r{\'e}nyi’s general theory of
  entropy,'' \emph{Autonomous Robots}, vol.~42, no.~2, pp. 235--256, 2018.

\bibitem{khosoussi14}
K.~Khosoussi, S.~Huang, and G.~Dissanayake, ``Novel insights into the impact of
  graph structure on slam,'' in \emph{2014 IEEE/RSJ International Conference on
  Intelligent Robots and Systems (IROS)}.\hskip 1em plus 0.5em minus
  0.4em\relax IEEE, 2014, pp. 2707--2714.

\bibitem{khosoussi19}
K.~Khosoussi, M.~Giamou, G.~S. Sukhatme, S.~Huang, G.~Dissanayake, and J.~P.
  How, ``Reliable graphs for slam,'' \emph{The International Journal of
  Robotics Research}, vol.~38, no. 2-3, pp. 260--298, 2019.

\bibitem{cheng81}
C.-S. Cheng, ``Maximizing the total number of spanning trees in a graph: two
  related problems in graph theory and optimum design theory,'' \emph{Journal
  of Combinatorial Theory, Series B}, vol.~31, no.~2, pp. 240--248, 1981.

\bibitem{chen20}
Y.~Chen, S.~Huang, and R.~Fitch, ``Active slam for mobile robots with area
  coverage and obstacle avoidance,'' \emph{IEEE/ASME Transactions on
  Mechatronics}, 2020.

\bibitem{kiefer74}
J.~Kiefer, ``General equivalence theory for optimum designs (approximate
  theory),'' \emph{The annals of Statistics}, pp. 849--879, 1974.

\bibitem{barfoot14}
T.~D. Barfoot and P.~T. Furgale, ``Associating uncertainty with
  three-dimensional poses for use in estimation problems,'' \emph{IEEE
  Transactions on Robotics}, vol.~30, no.~3, pp. 679--693, 2014.

\bibitem{brossard17}
M.~Brossard, S.~Bonnabel, and J.-P. Condomines, ``Unscented kalman filtering on
  lie groups,'' in \emph{2017 IEEE/RSJ International Conference on Intelligent
  Robots and Systems (IROS)}.\hskip 1em plus 0.5em minus 0.4em\relax IEEE,
  2017, pp. 2485--2491.

\bibitem{ganie19}
H.~A. Ganie, S.~Pirzada, R.~Ul~Shaban, and X.~Li, ``Upper bounds for the sum of
  laplacian eigenvalues of a graph and brouwer’s conjecture,'' \emph{Discrete
  Mathematics, Algorithms and Applications}, vol.~11, no.~02, p. 1950028, 2019.

\bibitem{de07}
N.~M.~M. De~Abreu, ``Old and new results on algebraic connectivity of graphs,''
  \emph{Linear algebra and its applications}, vol. 423, no.~1, pp. 53--73,
  2007.

\bibitem{zhu13}
Q.~Zhu and Y.~Wang, ``The kirchhoff index of weighted graphs,'' \emph{Int. J.
  Algebra}, vol.~7, pp. 267--280, 2013.

\bibitem{wald43}
A.~Wald, ``On the efficient design of statistical investigations,'' \emph{The
  annals of mathematical statistics}, vol.~14, no.~2, pp. 134--140, 1943.

\bibitem{carlone14}
L.~Carlone, R.~Aragues, J.~A. Castellanos, and B.~Bona, ``A fast and accurate
  approximation for planar pose graph optimization,'' \emph{The International
  Journal of Robotics Research}, vol.~33, no.~7, pp. 965--987, 2014.

\bibitem{carlone15}
L.~Carlone, R.~Tron, K.~Daniilidis, and F.~Dellaert, ``Initialization
  techniques for 3d slam: a survey on rotation estimation and its use in pose
  graph optimization,'' in \emph{2015 IEEE international conference on robotics
  and automation (ICRA)}.\hskip 1em plus 0.5em minus 0.4em\relax IEEE, 2015,
  pp. 4597--4604.

\bibitem{konolige10}
K.~Konolige, G.~Grisetti, R.~K{\"u}mmerle, W.~Burgard, B.~Limketkai, and
  R.~Vincent, ``Efficient sparse pose adjustment for 2d mapping,'' in
  \emph{2010 IEEE/RSJ International Conference on Intelligent Robots and
  Systems (IROS)}.\hskip 1em plus 0.5em minus 0.4em\relax IEEE, 2010, pp.
  22--29.

\bibitem{grisetti11}
G.~Grisetti, R.~K{\"u}mmerle, H.~Strasdat, and K.~Konolige, ``g2o: A general
  framework for (hyper) graph optimization,'' in \emph{Proceedings of the IEEE
  International Conference on Robotics and Automation (ICRA)}, 2011, pp. 9--13.

\bibitem{umari17}
H.~Umari and S.~Mukhopadhyay, ``Autonomous robotic exploration based on
  multiple rapidly-exploring randomized trees,'' in \emph{2017 IEEE/RSJ
  International Conference on Intelligent Robots and Systems (IROS)}.\hskip 1em
  plus 0.5em minus 0.4em\relax IEEE, 2017, pp. 1396--1402.

\end{thebibliography}
\end{document}